\documentclass[twoside,journal]{IEEEtran}

\usepackage{times}
\usepackage{epsfig}
\usepackage{graphicx}
\usepackage{amsmath}
\usepackage{amssymb}
\usepackage{multirow}
\usepackage{csquotes}
\usepackage{color,soul}
\usepackage{url}

\usepackage{xspace}

\usepackage[noadjust]{cite}
\renewcommand{\baselinestretch}{0.94}

\def\eg{\emph{e.g}.} 
\def\ie{\emph{i.e}.}

\def\etal{\emph{et al}.}

\usepackage[breaklinks=true,letterpaper=true,colorlinks,bookmarks=false]{hyperref}

\def\X{{\bf X}}

\newcommand{\rev}[1]{{\color{black}#1}}

\IEEEoverridecommandlockouts                              

\title{
Socially and Contextually Aware \\Human Motion and Pose Forecasting
}

\author{Vida Adeli$^{1}$, Ehsan Adeli$^{2}$, Ian Reid$^3$, Juan Carlos Niebles$^{2}$, Hamid Rezatofighi$^{2,3}$
\thanks{$^{1}$Faculty of Engineering, Ferdowsi University of Mashhad, Mashhad, Iran. {\tt\footnotesize vida.adeli@mail.um.ac.ir}}%
\thanks{$^{2}$Department of Computer Science, Stanford University, Stanford, CA 94035, USA {\tt\footnotesize {\{eadeli,\,jniebles\}@cs.stanford.edu}}}%
\thanks{$^{3}$School of Computer Science, University of Adelaide, Australia {\tt\footnotesize {\{hamid.rezatofighi,\,ian.reid\}@adelaide.edu.au}}}%
}

\begin{document}

\maketitle

\begin{abstract}
Smooth and seamless robot navigation while interacting with humans depends on predicting human movements. Forecasting such human dynamics often involves modeling human trajectories (global motion) or detailed body joint movements (local motion). Prior work typically tackled local and global human movements separately. In this paper, we propose a novel framework to tackle both tasks of human motion (or trajectory) and body skeleton pose forecasting in a unified end-to-end pipeline. To deal with this real-world problem, we consider incorporating both scene and social contexts, as critical clues for this prediction task, into our proposed framework. To this end, we first couple these two tasks by i) encoding their history using a shared Gated Recurrent Unit (GRU) encoder and ii) applying a metric as loss, which measures the source of errors in each task jointly as a single distance. Then, we incorporate the scene context by encoding a spatio-temporal representation of the video data. We also include social clues by generating a joint feature representation from motion and pose of all individuals from the scene using a social pooling layer. Finally, we use a GRU based decoder to forecast both motion and skeleton pose. We demonstrate that our proposed framework achieves a superior performance compared to several baselines on two social datasets.
\end{abstract}


\section{Introduction}
Forecasting human motion and body pose can be conducive in many real-world problems, \eg, in prediction of hazardous or anomalous behaviour for a smart surveillance system, in fine-grained anticipation of future activities, or in a collision avoidance system for an autonomous vehicle or a mobile robot navigating through a crowd.

The existing solutions for human skeleton pose forecasting focus on accurate prediction of intricate body joint movements as highly structured problem~\cite{martinez2017human,jain2016structural}. These approaches often concentrate on the problem of natural body joint prediction only while ignoring the global body motion reflecting these pose changes. For example, if a person is running, it is not only important to predict a skeleton pose representing this natural behaviour in future, it is also crucial to estimate the entire body displacement consistent with the predicted skeleton motions.
There exists also a parallel research thread focusing on forecasting human motion or trajectories~\cite{alahi2016social,gupta2018social}, by modeling it as prediction of a set of locations over time. However, this projected information may not be sufficient for the problems which require a detailed analysis of human body motions. 
Intuitively, the problem of human body pose forecasting and global motion prediction are inevitably correlated and should not be approached independently. In this paper, we aim to tackle these two problems in a unified framework.
\begin{figure}[!tbp]
  \centering
		\includegraphics[width=1.01\linewidth]{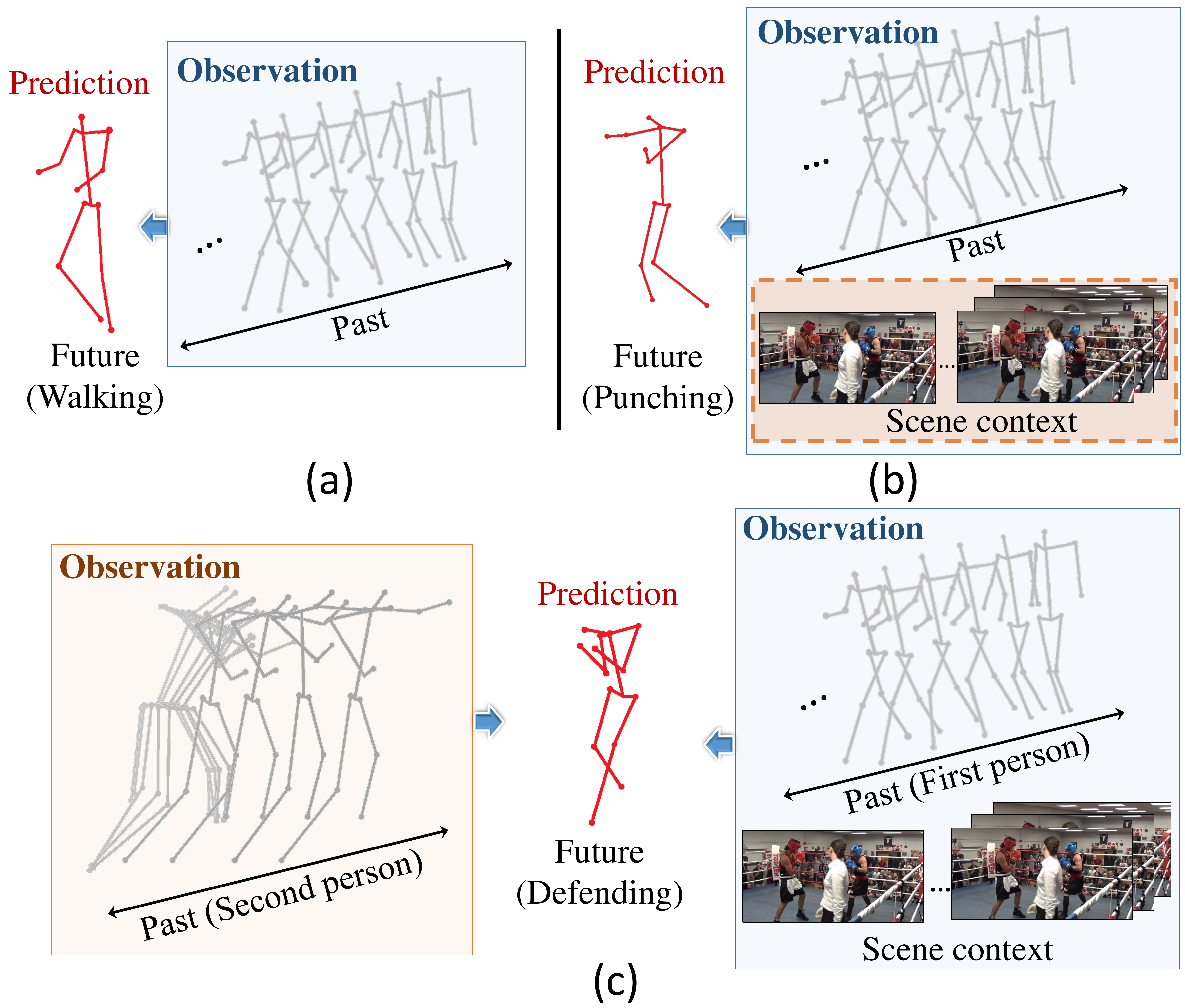}
	\caption{It is very challenging, even for human, to observe only a history of skeleton pose to predict the future pose and behaviour. In all three scenarios (a)-(c), the same motion and skeleton pose are provided as the observation, while in (b) the scene context (as a video) and in (c) both scene contexts (as a video) and social interactions (by observing the others' motion and skeleton pose) are additionally provided. It is intuitive both scene contexts and social clues are required to accurately predict the future human motion and pose. }
	\label{fig:teaser}
\end{figure}  
    
To accurately predict human skeleton pose and global motion, a model cannot solely rely on the history of body joints locations as input data. The physical context of the scene can provide important clues for this task. For example, considering an ice skating field, it is very likely that the body pose and motion of a person in this scene represent a sliding, but not diving or jogging,  activity. Therefore, incorporating this context as an informative clue into the pipeline would be helpful for tackling the problem. 

The social context in a scene can also provide valuable information for a pose and motion anticipating model.  In a real scene including individuals with different social connections and interactions, future gesture and pose of each person can be influenced by the other people activities and poses. For instance, consider a scene captured from a boxing game; if one of the boxers starts punching, it is predictable the opponent will hold a defending posture in near future (see Fig. \ref{fig:teaser}). Therefore, it is intuitive to build a framework which encodes these social interactions.

We aim to push the existing research thread in the community for human behavior prediction one step forward toward more practical scenarios, where human global motion and skeleton pose are jointly predicted while considering all social and contextual clues in the scene.  Our proposed framework jointly incorporates  a) the history of human trajectory and skeleton pose in the past using a recurrent based neural network encoder, \ie, Gated Recurrent Units (GRU) \cite{chung2015gated}, b) the physical context of the scene by encoding a spatio-temporal representation from the visual data, \ie, all past frames of a video stream, using the state-of-the-art I3D backbone \cite{carreira2017quo} and c) the social interaction between all individuals body and pose motion using a social pooling module. Furthermore, contrary to the majority of the prior works \cite{wang2019imitation,jain2016structural}, our model does not incorporate action labels during training and only builds one single action-agnostic model. We evaluate our proposed framework on two social datasets, created from NTU RGB+D 60~\cite{shahroudy2016ntu} and PoseTrack~\cite{andriluka2018posetrack}, and demonstrate its superior performance against several relevant baselines.  

In summary, the main contributions of the paper are:
\begin{enumerate}    	\setlength{\itemsep}{1pt}
	\setlength{\parskip}{0pt}
	\setlength{\parsep}{0pt}
    \item We propound the new problem of joint prediction of human \textbf{global} motion and body \textbf{pose} and introduce a proper evaluation metric (and also a loss) to measure the performance of both tasks in a single metric. 
    \item We propose an end-to-end learning framework for the task, which considers both \textbf{social} and visual \textbf{context} of the scene by incorporating justifiable learning modules such as a) GRU as a temporal encoder of skeleton and body motion, b) a social pooling layer (permutation invariant function) to generate a social feature, c) I3D backbone as context encoder and d) GRU decoder to forecast global motion and body skeleton pose.  
    \item Our framework achieves a \textbf{superior performance} in comparison with several relevant baselines and state-of-the-art models on two social datasets, including a 2D (PoseTrack \cite{andriluka2018posetrack}) and a 3D (NTU RGB+D 60 \cite{shahroudy2016ntu}) dataset.
\end{enumerate}
\section{Related work}
In this section, we review relevant literature on video based motion, trajectory, and pose forecasting. We also overview the prior work on social motion modeling to better position our proposed method and delineate its novelties.  

\subsection{Video-based forecasting} 
Prediction and forecasting of visual data based on videos was initially defined as extrapolating pixels values of the future frames   \cite{mathieu2016deep,zeng2017visual,hsieh2018learning,wang2018video,vondrick2016generating,mahjourian2017geometry}. Although these methods successfully predicted future frame(s) pixel values, it remains unclear how such a low-level future forecasting can be sufficient for analysis of events and scenes in the future. On the other hand, recent work focused on anticipating activities \cite{vondrick2016anticipating,kitani2012activity,sun2019relational,rhinehart2018first,soomro2018online,arzani2017structured,liang2019peeking} or trajectories \cite{alahi2014socially,deo2019scene,walker2015dense} (more details in the next subsection). Some other works have proposed predicting future semantic segmentation in videos \cite{luc2017predicting,rochan2018future,chiu2019segmenting}. In this work, we instead present a method that predicts the human dynamics, including global motion and local pose data. Such information can be used for detailed understanding of future human actions. 

\subsection{Motion and pose forecasting}
Human dynamics can be best modeled via global motion and detailed local joint locations (referred to as human skeleton or pose)  \cite{rogez2008randomized,sandriluka2010monocular,mehta2017vnect,chiu2019action,mangalam2019disentangling}. Human dynamics are previously modelled in images \cite{chao2017forecasting} by two major types of methods in videos including state transition models (such as graphical models) \cite{wang2008gaussian,wu2014leveraging} or more recently sequence-to-sequence deep learning methods \cite{ghosh2017learning,martinez2017human,jain2016structural,walker2017pose,barsoum2017hp,chiu2019action,wang2019imitation}. Chao \etal~\cite{chao2017forecasting} introduced a method for pose forecasting in 3D on static images, Barsoum \etal~\cite{barsoum2017hp} used Wasserstein GAN \cite{arjovsky2017wasserstein} in a probabilistic setting, Walker \etal~\cite{walker2017pose} used variational autoencoders,  Fragkiadaki \etal~\cite{fragkiadaki2015recurrent} proposed architectures based on LSTM and Encoder-Recurrent-Decoder methods, Yan \etal~\cite{yan2018mt} and Zhao \etal~\cite{zhou2018auto} proposed methods that could predict longer into the future, Martinez \etal~\cite{martinez2017human} introduced a designed RNN for human pose prediction, Chiu \etal~\cite{chiu2019action} utilized a multi-layer hierarchical recurrent architecture, and Wang \etal~ proposed to use imitation learning and specifically Generative Adversarial Imitation Learning (GAIL) \cite{Ho16} to capture human dynamics. However, all these works only predict the local dynamics of the pose data, \ie, they subtract global human motion from the joint coordinates and only predict these local movements. We argue that predicting both global and local motion information at the same time has better implications in terms of translating to real-world applications. Although a more challenging task, such modeling of the problem can leverage information from both global and local motion patterns and help better prediction of realistic 2D or 3D poses. Furthermore, all these previous works aim at only predict the joint coordinates of each single human in the scene. They ignore the social forces that other humans may deduce or neglect the context of the environment.

\subsection{Social models for motion predictions}
Even before the deep learning era, modeling human social interactions was a popular research topic among the community. Traditional models use hand-crafted features and rules such as ``social forces''  in order to model and predict human social interactions~\cite{helbing1995social,mehran2009abnormal,yamaguchi2011you,alahi2014socially,robicquet2016learning,pellegrini2009you}. However, the usability of these hand-crafted rules is restricted to their abstract level and the domain experts' information.   
Recently, due to the rise of popularity in development of autonomous driving systems and social robots and also with the recent developments in model-free data-driven frameworks for learning these social interactions, the problem of social human motion forecasting has received significant attention from both academia and industry. The main idea behind the modern socially-aware motion prediction approaches is to use a recurrent based neural networks along with a social pooling layer (a symmetrical function) to encode a social representation from this spatiotemporal data before forecasting the future path for each person~\cite{alahi2016social,lee2017desire,fernando2017soft+,gupta2018social, Sadeghian2018SoPhieAA,kosaraju2019social}. However the problem considered in these works is only about the task of social trajectory forecasting, \ie, prediction of a set of 2D locations over time.   
This simplified problem is lacking a detailed representation for this human body motions, which is crucial for human activity and behaviour anticipation. In a recent work, Joo \etal~\cite{joo2019towards} proposed a method for predicting social signals in triadic social interaction scenarios, in which the social signals (like facial expression and body position) of two other individuals are processed during the same time window, to estimate the same time frames signals of the target person.

In this paper, we aim to tackle this shortage by unifying the prediction of human motion and pose while incorporating social and scene contexts into the proposed framework.

\section{Social motion and pose forecasting model}
We define the task of motion and pose forecasting on a video with $T$ frames containing $N$ persons, where each person includes $L$ body joints. Note that we drop the index of the video for a better readability. 

Throughout the paper, we use $\X^j_o = \left(x^j_1, x^j_2, \ldots, x^j_t\right)$ to represent all observed values of body joint locations for $j^\text{th}$ person in a global coordinate up to time $t$, where $x^j_t\in\mathbb{R}^{L}$ is a vector of all body joint coordinates at each time.  Similarly, we use $\X^j_f = \left(x^j_{t+1}, x^j_{t+2}, \ldots, x^j_{t+T}\right)$ to demonstrate the future positions of these body joints, up to $T-(t+1)$ time-steps into the future.    



\begin{figure}[t]
\begin{center}
\includegraphics[width=\linewidth]{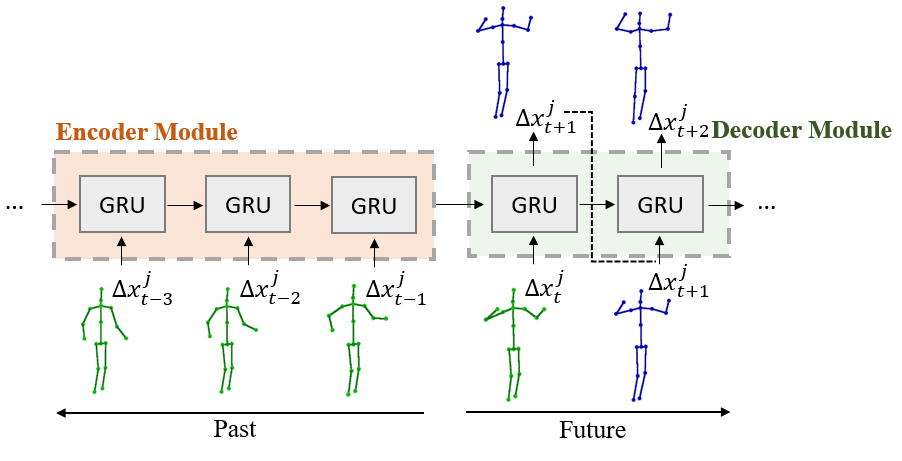}
\end{center}
  \caption{Sequence-to-sequence modeling of the human pose forecasting problem, using gated recurrent units. The network consists of an encoder module that encodes the historical (past) sequence of poses and a decoder module to recover the future poses. This modeling is for each human in the scene separately (\ie, it is blind to social or context-related forces).}
\label{fig:arch1}
\end{figure}

\begin{figure*}
\begin{center}
\includegraphics[width=\linewidth]{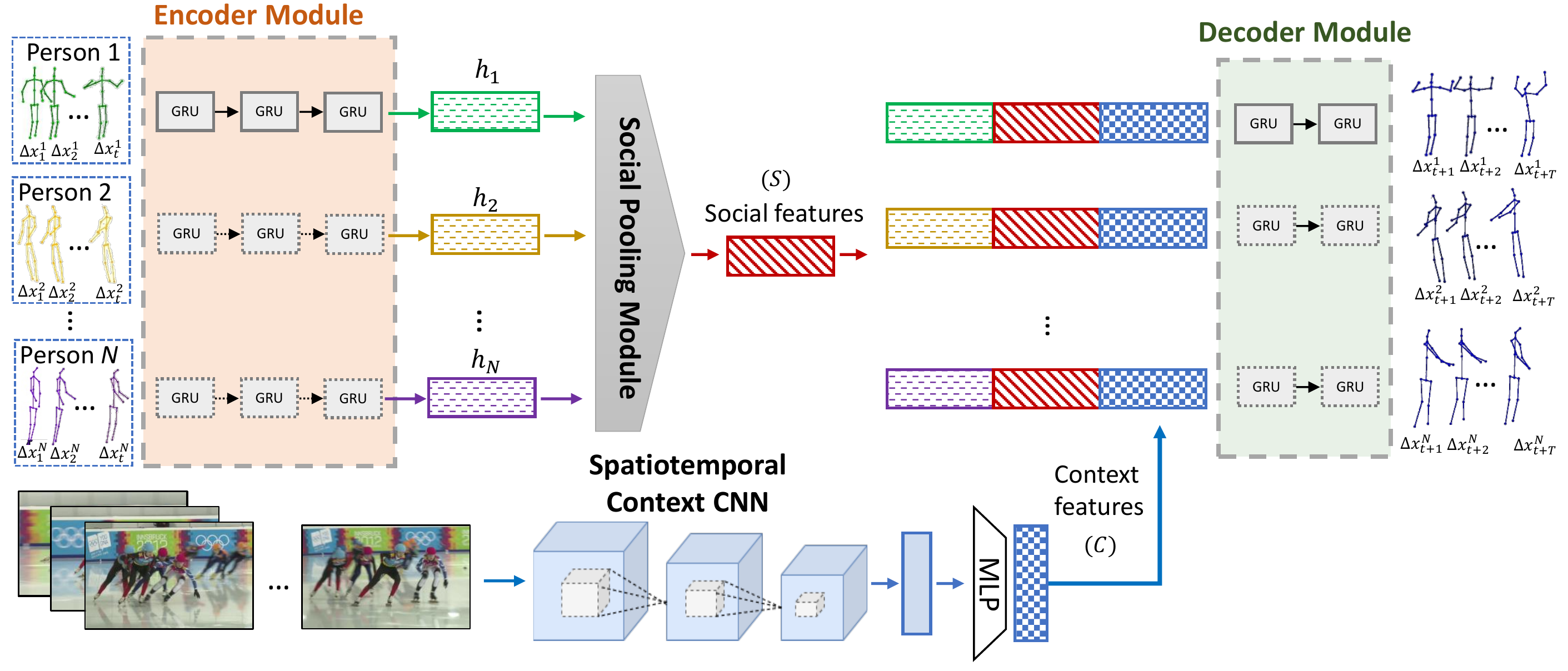}
\end{center}
  \caption{Overview of the proposed socially and contextually aware human motion and pose forecasting model. Our model consists of four main modules: (1) \textbf{Encoder Module} encodes the historical (past) sequence of poses for each individual in the scene; (2) \textbf{Social Pooling Module} produces one shared socially aware feature vector by pooling (a permutation invariant operation) features across all embeddings; (3) \textbf{Spatiotemporal Context Encoding Module} uses an I3d model to extract spatiotemporal features from the videos, which are passed through two layers of fully connected neural network, MLP, to create the shared context features; (4) \textbf{Decoder Module} receives the concatenated person-specific embedding, shared social, and context features to produce the predicted poses for each future time step. }
\label{fig:arch2}
\end{figure*}

\subsection{Forecasting motion and pose jointly}
The dominant trend for forecasting the human pose is to predict the future position of body joints relative to the body center (neck or torso)~\cite{jain2016structural,chiu2019action,wang2019imitation,martinez2017human}. Intuitively, it is an easier task for a model to only learn the body joint movements respected to a centered body skeleton, making the predicted values (as offsets), zero mean variables. Similarly, with the same logic, the global body motion is often modelled as an offset prediction task from the last predicted position instead of predicting their absolute coordinates~\cite{Sadeghian2018SoPhieAA}. To forecast both components jointly, one trivial extension is to estimate an offset value for each component independently, \ie, one offset for the entire body motion and one set of offsets for each body joints motion relative to the body. However, there are two potential problems with this approach: I) this setting will make the behaviour of joints and body motions uncorrelated since their values are predicted independently, and II) the evaluation and ranking of overall task will become complicated as each component has its own error. To address this, we first suggest a very simple, but an intuitive, metric which encodes both source of errors jointly by a single value. To this end, we consider the real-world setting of movements of joints in space and time in \emph{the original space}. Therefore, we define the evaluation metric as  a norm error, \eg, the MSE or $\ell_2$ error, of the ground-truth ($\X_f$) and predicted ($\hat{\X}_f$) absolute future locations of the joints with respect to \emph{their global coordinate}, \eg, $\text{MSE}(\X_f,\hat{\X}_f)$. This metric encodes both global body motion and joint movements respect to the body jointly. However, for the loss as offset prediction is easier task, we minimize the error between the offsets values of the predicted joint locations and the ground-truth joint locations, \ie, $\mathcal{L}= \mathbb{E}_{j} \left(\text{MSE}(\Delta\X^j_f,\Delta\hat{\X}^j_f)\right)$.



In order to encode and decode this temporal data, we apply sequence-to-sequence model. To this end, we use a previous state-of-the-art recurrent model~\cite{martinez2017human} with GRU cells to implement the encoder and decoder modules. In~\cite{martinez2017human}, the encoder receives the past pose states (as velocities/offsets of joints movements with respect to the body location) up to time $t$ and the decoder outputs the predicted pose states for the time steps $t+1$ to $T$ (Fig.~\ref{fig:arch1}). 

\rev{Our framework uses exactly the same encoder and decoder architecture, \emph{i.e.} identical number of parameters and layers, as~\cite{martinez2017human}}; However as shown in Fig.~\ref{fig:arch2}, we feed the encoder for each person with the person's body and pose motion, $\X_o^j$, to generate a person-specific representation of the historical (past) pose and motions, denoted by $h^j\in\mathbb{R}^K$, where $K$ is the dimension of the feature. These embedding representations are shown by colored and textured boxes, \eg, green, orange, and purple, immediately after encoder module in Fig.~\ref{fig:arch2}. Then, this representation is concatenated with social and context features before being fed into the decoder, in order to forecast body motion and pose as $\Delta\X^j_f$.   

\subsection{Social and context modules}

To incorporate social interaction among the people in the scene, we need to encode a collective feature representation from everyone's body motion and pose. We can use their encoded features, $h^j$. This feature cannot be simply attained by concatenation of individuals encoded features, $h^j$, as this representation should not be sensitive to the way each person is ordered. Any ordering of the same set of instance input should generate an identical feature representation. To design this module, similar to all set encoding problems, \eg, point clouds~\cite{zaheer2017deep,qi2017pointnet} and social trajectories~\cite{alahi2014socially,gupta2018social}, we need a symmetrical (permutation invariant) function, $\mathcal{H(\cdot)}$, applied on all $h^j$ embeddings (as a set) to generate a permutation invariant joint representation $S\in\mathbb{R}^K$, \ie, 
\begin{equation}
    S = \mathcal{H}(\{h^1,\cdots,h^j, \cdots, h^N\}), 
\end{equation}

The family of these symmetric functions is vast and many mathematical functions has this symmetrical property. However, the most popular ones for being incorporated easily into deep learning pipelines are \textit{max}, \textit{sum} and \textit{average} pooling layers used in~\cite{zaheer2017deep,qi2017pointnet,alahi2014socially,gupta2018social}. We explore the choice of some of these functions in our experiments (see Table~\ref{tlb:poolingstrategies}). Then, the pooled social feature, $S$, denoted by the red boxes in Fig.~\ref{fig:arch2} is concatenated with features of each person, $h^j$. 



To extract the context features, we encode a spatio-temporal representation from the video. To this end, we use a pre-trained I3D model~\cite{carreira2017quo}, applied on the RGB video, from which the individual poses were extracted. This leads to another embedding vector that goes through two layers of fully connected neural networks (denoted by multi-layer perceptron, MLP, in Fig.~\ref{fig:arch2}). The resulting context feature vector, denoted by $C$ (shown with a blue box in Fig.~\ref{fig:arch2}), are additional features shared across all persons in the scene.

Finally, the decoder module for each person $j$ receives a socially and contextually aware feature vector, generated by the concatenation of person specific embedding $h^j$, the shared socially pooled features $S$, and the shared context features $C$. The decoder outputs pose, with both global and local movements encoded in it. 
\section{Experiments}
    In this section we present the experimental results on two challenging datasets: NTU RGB-D 120 ~\cite{shahroudy2016ntu} and PoseTrack ~\cite{andriluka2018posetrack}. Different sets of experiments are conducted to assess the impact of our contributions and  finally, the results are compared with different baseline methods.
    
    \noindent\textbf{Evaluation metric.}
    As discussed earlier and consistent with the prior work ~\cite{martinez2017human,chiu2019segmenting,wang2019imitation}, we use MSE metric, which is the $\ell_2$ distance between the ground-truth and the predicted poses at each time step $i \in \{t+1, \ldots, T\}$ but averaged over the number of persons $N$ in that frame. Contrary to the previous work that center the poses (\ie, remove the global body motion from joints motion), our metric evaluates the differences between the predictions and the ground-truth in \emph{the original global space}. 

    \begin{table*}
    \small
    \begin{center}
    \caption{\rev{Error (MSE) on \textbf{PoseTrack} (in pixels)  and \textbf{NTU RGB+D} (in cm) datasets for different baselines and our proposed models. In each column, the best obtained result is highlighted with boldface typesetting and the second best is underlined. In the first eleven experiments, the first word before the hyphen notation, is an indicator of method used for body skeleton pose and the second one is for method used as global motion (trajectory). \textbf{Zero}: Zero velocity, \textbf{Constant}: Constant velocity, \textbf{LocalPose}: The results of the model trained on the centered poses whether by the original model ~\cite{martinez2017human} or using our social module, \textbf{SLSTM}: The results for learning global motion using Social-LSTM~\cite{alahi2016social}, \textbf{SGAN}: The results for learning global motion using Social-GAN~\cite{gupta2018social}.}}    \label{tlb:posetrackandNTU}
     \setlength{\tabcolsep}{7pt}
    \begin{tabular}{|l|c c c c c||c c c c|}
    \hline
    & \multicolumn{5}{c||}{PoseTrack} &\multicolumn{4}{c|}{NTU RGB + D}\\ \hline
    \multirow{2}{*}{Method}& \multicolumn{5}{c||}{milliseconds} &\multicolumn{4}{c|}{milliseconds}\\
    &80&160&320&400&560&80&160&320&400\\
    \hline\hline
    ZeroPose-ZeroMotion & 153.3 & 263.7 & 432.3 & 473.9 & 563.5 & 14.1 & 22.1 & 34.8 & 40.5\\
    ZeroPose-ConstantMotion & 154.0 & 265.4 & 436.9 & 479.6 & 572.6 &14.2 & 23.0 & 35.7 & 44.2\\
    ConstantPose-ConstantMotion & 154.4& 266.6 & 440.4 & 484.3 & 579.5&15.3 & 23.6 & 36.1& 46.8\\\hline\hline
    LocalPose \cite{martinez2017human}-ZeroMotion & 72.4& 114.7 & 217.8 &253.1&311.5& 14.0 & 23.6 & 34.9 & 40.3\\
    LocalPose \cite{martinez2017human}-ConstantMotion& 68.3 & 109.7 & 206.9 & 249.4 & 306.6& 14.1 & 24.0 & 34.8 & 43.1\\
    LocalPose (+ Social pooling)-ZeroMotion & 52.2& 98.9 & 186.6 & 228.3 & 282.9& 14.0 & 22.0 & 34.1 & 40.1\\
    LocalPose (+ Social pooling)-ConstantMotion & 47.2 & 90.2 & 173.5 & 207.7 & 274.6& 13.9 & 21.8 & 33.8 & 39.2\\\hline\hline
    \rev {LocalPose \cite{martinez2017human}-SGAN (social) \cite{gupta2018social}}\ & \rev{62.1} & \rev{105.6} & \rev{189.9} &\rev{230.8}&\rev{282.6} &\rev{14.9} & \rev{24.7} & \rev {36.1} & \rev {43.9}\\
    \rev {LocalPose \cite{martinez2017human}-SLSTM (social) \cite{alahi2016social}}\ & \rev{63.1} & \rev{107.3} & \rev{193.4} &\rev{235.4}&\rev{285.7} &\rev{14.2} & \rev{24.1} & \rev {35.7} & \rev {43.5}\\
    \rev {LocalPose (+ Social pooling)-SGAN (social) \cite{gupta2018social}}& \rev{45.6} & \rev{84.2} & \rev{167.3} &\rev{203.9}&\rev{269.6}& \rev{14.5} & \rev{22.4} & \rev {34.7} & \rev {40.9}\\
    \rev {LocalPose (+ Social pooling)-SLSTM (social) \cite{alahi2016social}}& \rev{45.9} & \rev{84.8} & \rev{168.2} &\rev{204.1}&\rev{268.3}& \rev{14.3} & \rev{22.3} & \rev {34.5} & \rev {40.5}\\\hline\hline
    \textbf{Ours (JointLearning)} & \textbf{43} & 81.7 & 158.1 & 197.2 & 256.3& 13.1 & 20.0 & 30.5 & 35.1\\
    \textbf{Ours (JointLearning+Context)} & 44.2 & 82.1 & 157 & 194.2 & 251.5& \underline{13.0} & 20.1 & 30.4 & 35.0\\
    \textbf{Ours (JointLearning+Social)} & \underline{43.1} & \underline{81} & \underline{154.3} & \underline{191} & \underline{246.7}& \textbf{12.8} & \textbf{19.6} & \underline{29.9} & \underline{34.3}\\
    \textbf{Ours (JointLearning+Social+Context)} & \textbf{43} & \textbf{80.1} & \textbf{152.3} & \textbf{188.4} & \textbf{243.2}& 13.1 & \underline{19.8} & \textbf{29.7} & \textbf{34.1}\\
    \hline
    \end{tabular}
    \end{center}
    \vspace{-5mm}
    \end{table*}

\subsection{Experimental settings}
    We use a sequence-to-sequence architecture as our encoder-decoder model with GRU modules, exactly the same as \cite{martinez2017human}, as the RNN for both encoder and decoder. We consider a fixed hidden state dimension of 1024 for GRU modules in the encoder and for decoder, it varies depending on the selected dimension for the context feature after embedding to the MLP. We use a two layer dense network for the context features with a dropout probability of 0.7 that reduces its dimension to 256, before feeding it to the pose and motion forecasting model.
    
    The I3D model~\cite{carreira2017quo} pre-trained on Kinetics \cite{kay2017kinetics} is used as the context network to extract the spatio-temporal context features from video frames. The I3D 1024-dimensional feature vectors are then used as the context features in our framework. The hyper-parameters are selected through experiments on a validation set. We used a learning rate with initial value of $5e^{-4}$ and a learning rate decay factor of 0.95 for training the model with the Adam optimizer. For all the experiments, the training is performed over all activity types resulting in a single action agnostic model. The best model is selected with early stopping on the validation set.
    
\subsection{Datasets}
    There are several commonly used benchmark datasets for human pose forecasting such as Human 3.6M ~\cite{ionescu2013human3} and Penn action dataset ~\cite{zhang2013actemes}. However, these  publicly available datasets being used by the previous work ~\cite{liu2019towards, martinez2017human, chiu2019action, fragkiadaki2015recurrent, walker2017pose} focus on single isolated individuals and do not entail social interactions. Moreover, most of these datasets are short-length video sequences, that makes it hard to encode the social behaviour of individuals. To this end, to test the performance of our model for social human pose forecasting, we run extensive experiments on NTU RGB+D 60 and PoseTrack datasets that consist of multiple persons interacting with each other to complete different actions.
    
    \noindent\textbf{PoseTrack:} The PoseTrack is a large-scale multi-person dataset based on the MPII Multi-Person benchmark. The dataset covers a diverse variety of interactions including person-person and person-object in dynamic crowded scenarios. Pose annotations are provided for 30 consecutive frames centered in the middle of the sequence. The pose forecasting in this dataset is challenging because of the wide variety of human actions in real-world scenarios and the large number of individuals in each sequences with large body motions and occlusions. In each sequence, we select those individuals that are present in all frames.
    For experimenting on this dataset, we trained our model by observing the past 15 frames as the history of each person and forecast the future 15 frames. We use a set of 14 joints in 2D space as the human pose, including the head, neck, shoulders, elbows, wrists, knees, hips, and ankles. The data being used is in image coordinate and therefore the results are reported in pixel. Also, we use 60\% of sequences in training split for training our model. The rest were split equally for validation and test.
    
    \noindent\textbf{NTU RGB+D 60:} The NTU dataset contains both single-person actions and mutual actions. We selected mutual sequences for our experiments, which are in 11 different action categories including \textit{punching}, \textit{kicking}, \textit{pushing}, \textit{pat on back}, \textit{point finger}, \textit{hugging}, \textit{giving object}, \textit{touch pocket}, \textit{handshaking}, \textit{walking toward}, and finally \textit{walking apart} actions. 
    In this dataset, each pose is represented by 3D locations of 25 major body joints at each frame. Since some of the joints, such as finger positions, are too fine-grained and are not important in our problem, we use only 13 body joints for our experiments. The pose locations are in camera coordinate and hence the results are reported in centimeter (cm). As suggested in \cite{shahroudy2016ntu}, we utilize the standard cross-subject evaluation in which the 40 subjects are split into different groups of training and testing. Also, for validation set we divided the test set into two splits. The model is trained by observing poses in 40 frames and then forecasts for the next 10 frames. Finally, in contrast with previous work \cite{fragkiadaki2015recurrent, martinez2017human, jain2016structural} that propose action-specific models or models reliant on semantic knowledge about action categories, we train a single action agnostic model that does not require any prior knowledge about the action categories.

\subsection{Baselines}
    Since no other work consider the problem of social pose forecasting,
    there is no previously published work on social datasets. As we are considering two concepts of pose and global motion, to show the effectiveness of modeling them jointly, we conduct some experiments separating these concepts and compare our method against following baselines. \rev{Three sets of baseline experiments are conducted (The first three sets in Table \ref{tlb:posetrackandNTU}. In the first set of baselines no learning is applied in the method, whereas in the second set, pose information is trained while there is no learning for the global motion. These baselines provide of with a sanity check of our data and show how well can a trivial baseline with no learning perform on our datasets. For the third set of experiments both pose and global motion are learned but separately, using the state-of-the-art methods. Then, in all the baselines, the predicted global motion is added to the predicted pose for estimating metric values in every time step. As the final set of experiments, our method jointly learns and predicts the global motion and pose, involving the social interactions and context information.}
      
    \noindent \textbf{Zero Pose and Zero Motion velocity}, denoted as \textit{ZeroPose-ZeroMotion},  assumes that all the future predictions are identical to the very last seen pose and thus outputs the last observed pose for the future in a same location with no global motion. In other words, pose at time $t$ remains with no change in joint and displacement. The zero velocity baseline for pose has been used by many other methods for comparison and demonstrated to be a high performance baseline that is hard to outperform ~\cite{fragkiadaki2015recurrent, martinez2017human, toyer2017human}. 
    
    \noindent \textbf{Zero Pose and Constant Motion velocity}, denoted as \textit{ZeroPose-ConstantMotion}, considers that in future prediction the joints relative position remains fixed and the whole body moves with a constant velocity.
    
    \noindent \textbf{Constant Pose and Constant Motion velocity}, denoted as \textit{ConstantPose-ConstantMotion}, assumes that both pose and global trajectory move with a constant velocity model.
      
    \noindent \textbf{Local pose forecasting with zero/constant global trajectory}. In order to show that it is fundamental to model both the movements of the joints and the global trajectory jointly, we experiment another baseline that uses the centered human poses as its input, just like almost all previous pose forecasting methods that exclude global body motion. To do so, we center the poses by subtracting the neck position and train a model to predict centered poses. Then during prediction, we add the zero or constant velocity to create the global trajectory. We report the results by training the local pose  (i) using the vanila model proposed in \cite{martinez2017human}, or (ii) by incorporating our social pooling module into this model~\cite{martinez2017human}.
    
    \noindent \rev {\textbf{Local pose forecasting with SGAN/SLSTM global trajectory forecasting}. In this set of baselines, pose information is trained exactly in the same manner as previous set (centered with vanila and social model), while the global motion is also learned by two state-of-the-art methods in trajectory prediction which are \textit{Social GAN} \cite{gupta2018social} and \textit{Social LSTM} \cite{alahi2016social}. These baselines can clearly show the difference between performance of forecasting while both pose and trajectory being trained seperately or jointly.}
    
    \noindent \textbf{Our joint learning model (with or without context)} baseline uses exactly the same model as \cite{martinez2017human} followed by our proposed loss for learning to predict body pose and global motion, jointly. For our joint learning model (i.e., non-social model) with context, we also integrate the context feature before decoding the future output. However, these baselines do not contain the social module. 
    
    \noindent \textbf{Our joint learning social model without context}. In this baseline we intend to investigate the effect of scene context on the proposed framework. The model in this baseline excludes the spatiotemporal context encoding module of our final model and the other modules are remained exactly the same. 
     

    \begin{table}
    \small
    \begin{center}
    \caption{Effect of different social pooling strategies in our social+context motion and pose forecasting model. }    \label{tlb:poolingstrategies}

    \begin{tabular}{|l|p{0.6cm}p{0.6cm}p{0.6cm}p{0.7cm}p{0.7cm}|}
    \hline
    \multirow{2}{*}{Method}& \multicolumn{5}{c|}{milliseconds} \\
    &80&160&320&400&560\\
    \hline\hline
    Average Pooling & 43.9 & 81.3 & 153.6 & 190.5 & 246.8\\
    Sum Pooling &44.1&81.9&154.1&190.9&247.2\\
    \textbf{Max Pooling} &  \textbf{43} & \textbf{80.1} & \textbf{152.3} & \textbf{188.4} & \textbf{243.2}\\
    \hline
    \end{tabular}
    \end{center}
    \vspace{-5mm}
    \end{table}

    \begin{table*}
    \small
    \begin{center}
    \caption{MSE action level errors (in cm) on the 11 mutual action categories of NTU RGB+D 60 dataset. \rev{Ours (JointLearning) is the extended version of \cite{martinez2017human} which uses the local pose and global motion jointly It does not consider the social term.}}     
    \label{tlb:actionlevel}

     \vspace{-2.5mm}
    \begin{tabular}{|l|p{0.6cm} p{0.6cm} p{0.6cm} p{0.6cm}|p{0.6cm} p{0.6cm} p{0.6cm} p{0.6cm} |p{0.6cm} p{0.6cm} p{0.6cm} p{0.6cm}|}
    \hline
     & \multicolumn{4}{c|}{Punching} & \multicolumn{4}{c|}{Kicking} & \multicolumn{4}{c|}{Pushing} \\\hline
     milliseconds &80&160&320&400&80&160&320&400&80&160&320&400\\ \hline\hline
     Ours (JointLearning)&15.7&24.5&38.3&44.08&16.9&28.1&45.9&53.3&19.7&30.8&52.9&61.7\\ 
    Ours (JointLearning+Social+Context)&\textbf{15.6}&\textbf{22.4}&\textbf{37.8}&\textbf{44.1}&\textbf{16.7}&\textbf{27.8}&\textbf{44.9}&\textbf{52.1}&\textbf{19.3}&\textbf{29.8}&\textbf{50.4}&\textbf{58.4}\\\hline\hline
    
    & \multicolumn{4}{c|}{Pat on back}& \multicolumn{4}{c|}{Point finger} & \multicolumn{4}{c|}{Hugging}\\\hline
    milliseconds &80&160&320&400&80&160&320&400&80&160&320&400\\ \hline\hline
    Ours (JointLearning)&\textbf{5.4}&\textbf{7.9}&\textbf{10.7}&\textbf{12.2}&\textbf{5.9}&\textbf{8.2}&\textbf{10.2}&\textbf{12.0}&25.6&36.9&51.3&56.6\\ 
    Ours (JointLearning+Social+Context)&5.5&\textbf{7.9}&10.8&12.3&\textbf{5.9}&8.4&10.7&12.6&\textbf{25.5}&\textbf{36.7}&\textbf{50.1}&\textbf{54.9}\\\hline\hline
    
    & \multicolumn{4}{c|}{Giving object} & \multicolumn{4}{c|}{Touch pocket} & \multicolumn{4}{c|}{Handshaking}\\\hline
    milliseconds &80&160&320&400&80&160&320&400&80&160&320&400\\ \hline\hline
    Ours (JointLearning)&10.2& 14.5&22.0&26.3&7.1&10.8&17.1&20.1&\textbf{8.0}&\textbf{11.1}&\textbf{14.5}&\textbf{16.0}\\ 
    Ours (JointLearning+Social+Context)&\textbf{10.1}&\textbf{14.4}&\textbf{21.6}&\textbf{25.7}&\textbf{7.0}&\textbf{10.6}&\textbf{16.5}&\textbf{19.0}&\textbf{8.0}&11.2&14.9&16.5\\\hline\hline
    
    & \multicolumn{4}{c|}{Walking towards} & \multicolumn{4}{c|}{Walking apart} & \multicolumn{4}{c|}{Average of all 11} \\\hline
    milliseconds &80&160&320&400&80&160&320&400&80&160&320&400\\ \hline\hline
    Ours (JointLearning)&23.6&35.7&56.3&62.9&21.8&36.6&58.7&68.6&13.1&20.0&30.5&35.1\\ 
    Ours (JointLearning+Social+Context)&\textbf{23.0}&\textbf{35.0}&\textbf{54.9}&\textbf{61.3}&\textbf{21.4}&\textbf{35.7}&\textbf{57.2}&\textbf{66.7}&\textbf{13.1}&\textbf{19.8}&\textbf{29.7}&\textbf{34.1}\\
    \hline
    
    \end{tabular}
    \end{center}
    \vspace{-5mm}
    \end{table*}

    \begin{figure}[t]
    \begin{center}
    \includegraphics[width=0.5\textwidth]{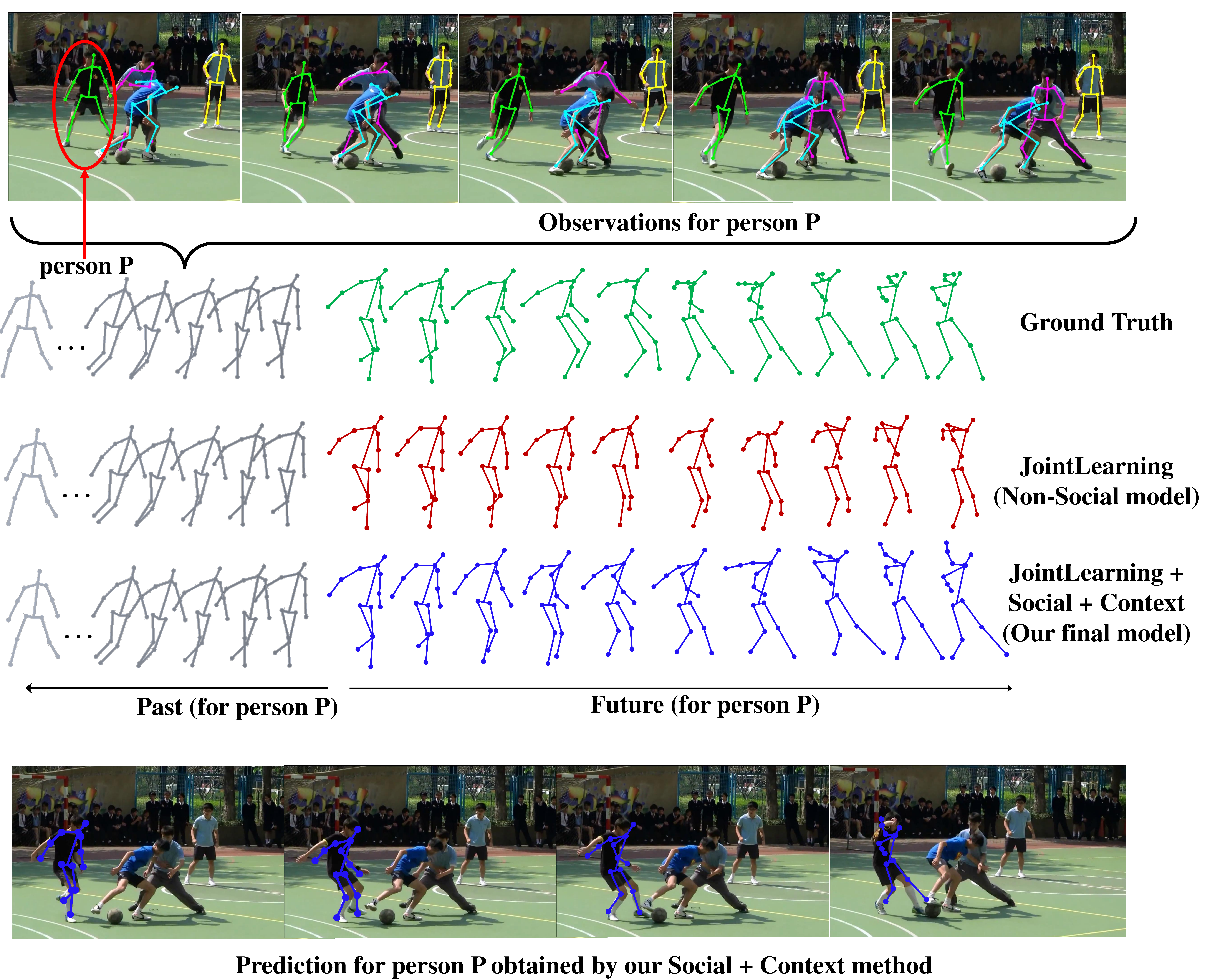}
    \end{center} \vspace{-15pt}
      \caption{Visualization of the first 10 pose predictions for a sample sequence from the action playing football in PoseTrack dataset. The first row contains a number of sample observations that the model receives as input including the history of all persons in the scene. The next three rows show historical and future poses for Person P (circled out in the top row). The gray poses are ground-truth data of the history of action. The green, red and blue poses are the future poses of ground-truth, JointLearning Non-social method and our final method (JointLearning+Social+Context), respectively. The last row shows the predicted poses by our social+context model on the frames.}
    \label{fig:fig4}
    \end{figure}
    
\subsection{Comparison with the baselines}
    We compare our proposed method against different baselines on both PoseTrack and NTU RGB+D datasets in Table \ref{tlb:posetrackandNTU}. It is important to note that in all baselines, the results are reported as the previously described metric in global mode. In the first set of experiments, denoted as \textit{ZeroPose-ZeroMotion}, \textit{ZeroPose-ConstantMotion}, and \textit{ConstantPose-ConstantMotion}, we considered the pose and global trajectory separately and there is no learning involved in these baselines. The reason that the zero or constant baselines are performing more poorly in the PoseTrack dataset than in the NTU is that the PoseTrack contains more complex and realistic scenarios with severe body motions in comparison to the NTU dataset that is recorded in a controlled laboratory settings. Quantitative evaluations on NTU dataset in previous studies~\cite{fragkiadaki2015recurrent} confirmed this fact that most existing methods could be outperformed by zero-velocity and constant velocity baselines. However, by considering both body pose and global motion jointly, even without incorporating the social and scene context information, we could outperform these baselines in both datasets. Even the vanila LocalPose model results in better predictions in PoseTrack. However, this superiority for the vanila LocalPose model is marginal in NTU due to the aforementioned limitations in this dataset. 
    
    In the second set of experiments, denoted as \textit{LocalPose-(Zero/Constant) Motion}, the pose is centered and the model learns to predict a centred pose \cite{martinez2017human}. Then, we place the predicted poses in the position of time $t$ (zero velocity on the neck) or we move the whole body with constant velocity. As can be seen, modeling both pose and global motion \rev{(Our joint learning results)} outperforms each of these methods. This shows that adding a constant velocity model to the centered pose model cannot compensate the global motion term and so they are highly correlated to each other and should be modeled jointly. \rev{In the third set of experiments both person centric pose and global motion are learned by state-of-the-art pose forecasting (non-social \cite{martinez2017human} and social) and trajectory forecasting models (SGAN \cite{gupta2018social} and SLSTM \cite{alahi2016social}) separately and the final global pose is obtained by adding the global motion prediction to the predicted local pose. These experiments also demonstrates that pose is better predicted when jointly predicting the global motion and pose. Inspecting precisely, we can observe that methods with constant or zero motion do not considerably differ from trajectory learnable methods in terms of results, in NTU dataset. This is due to small amount of global motion in NTU, which is recorded in constraint laboratory settings. Even in some cases, we see that the complexity of the trajectory model (SGAN/SLSTM) results in worse predictions. Despite this, the results show that the use of joint learning with social and context terms improved the final results in NTU.}

    Finally, as the forth set of experiments, we consider our joint learning (non-social) model, which is an extension of the state-of-the-art method \cite{martinez2017human} but with our new loss defined on both local pose and global motion jointly. Then, we add the context and social module in order to investigate the effects of each the social and context information in our model. As can be seen, by using the social and context information of the scene and modeling the pose and global motion jointly, our model achieves the highest performance on both NTU and real-world PoseTrack dataset. The results indicate that not only utilizing the social interactions, but also the context information could improve the prediction result. This claim is supported by the lower prediction errors of Social + Context model compared to the Social method. We believe the reason that the context make less improvement on NTU is the constraint environmental settings that the videos are recorded in. All videos are obtained in a laboratory setting and therefore concepts of context become impractical in that condition.

    Finally, we investigated the effect of different social pooling strategies in the proposed Social + Context pose and motion forecasting models (Table \ref{tlb:poolingstrategies}). The best results are achieved using \emph{max} pooling strategy.

    In Fig. \ref{fig:fig4}, we show a sample qualitative result from the action playing football, obtained by our final (JointLearning+Social+Context) model against the results for our JointLearning (Non-social) model (JointLearning), which is the extended version of \cite{martinez2017human} with loss jointly on local pose and global motion. As it can be seen in this figure, compared to JointLearning Non-Social model the predictions from our final model (blue poses) are more visually closer to the ground-truth poses (green skeletons). This improvement in predictions increase over time, when the social interaction becomes more meaningful. For example, when a person observes the tackling action by another person, he would have some specific pose or change in direction of motion to avoid it. Besides, observing the scene context as another source of information has a great impact on the final result. Because when the model observes the football pitch, it can better predict the movements of leg joints. Moreover, this claim is also supported by results in Table \ref{tlb:actionlevel}, in which we can see our final model (JointLearning+Social+Context)  outperforms the Non-social model (JointLearning) in most of the action classes. Precisely investigating the results in this table, we can figure out that in those actions that involve a high amount of social interaction such as ``punching" or ``hugging", our ultimate  model (JointLearning+Social + Context) outperforms the JointLearning Non-social method and only in a few number of actions with minor social interactions the JointLearning Non-social method performs better. On average, our final model results in better predictions in all time steps and this improvement increases over time. 
    
    Finally, (Table \ref{tlb:jointloss}) shows that pose is better predicted when jointly predicting the global motion and pose with a joint loss rather than separating the two information and learning them separately (Human-centric pose and motion). This experiment is conducted on the single-person subset of NTU RGB-D 60 dataset. For the first experiment the pose is centered relative to the neck and the value of neck is concatenated to the input vector as the human motion.
    
    \begin{table}
    \small
    \begin{center}
    \caption{MSE errors (in cm) on the model trained on 1) separated pose and motion and 2) a global pose with joint loss.}    \label{tlb:jointloss}
     \vspace{-2.5mm}
     \setlength{\tabcolsep}{4pt}
    \begin{tabular}{|p{5cm}|p{0.6cm}p{0.6cm}p{0.6cm}p{0.7cm}|}
    \hline
    \multirow{2}{*}{Metric (MSE)}& \multicolumn{4}{c|}{milliseconds} \\
    &80&160&320&400\\
    \hline\hline
    Summed over local pose and trajectory  & 16.7 & 18.7 &20.0 & 24.0 \\
    {On global pose with joint loss} &  \textbf{16.4} & \textbf{18.4} & \textbf{19.7} & \textbf{23.8}\\
    \hline
    \end{tabular}
    \end{center}
    \vspace{-5mm}
    \end{table}

\section{Conclusion}
In this paper, we proposed an action agnostic model for simultaneously forecasting global motion (trajectory) and local pose movements. Our model incorporates social and context cues for making the predictions. Contrary to the previous work, we defined a metric and a loss function in the original space of the movements (2D in PoseTrack and 3D in NTU RGB+D) without the global motion removed. Our end-to-end training framework based on encoder-decoder GRU cells outperforms all the appropriate baselines and shows that the use of spatiotemporal context and social considerations both improve the global pose prediction performance. Such modeling can facilitate detailed understanding of the future actions, which can be used for better navigation and human-robot interaction. As a direction for the future work, GAN-based social models can improve the results.

\noindent\textbf{Acknowledgement:} 
This research was partially supported by Mindtree, Oppo, and Panasonic.

{\small
\bibliographystyle{ieeetr}
\bibliography{refs}
}

\end{document}